# Performance Study of YOLOv5 and Faster R-CNN for Autonomous Navigation around Non-Cooperative Targets

Trupti Mahendrakar
Florida Institute of Technology
150 W. University Blvd
Melbourne, Fl 32901
tmahendrakar2020@my.fit.edu

Andrew Ekblad
Florida Institute of Technology
150 W. University Blvd
Melbourne, Fl 32901
aekblad2019@my.fit.edu

Nathan Fischer
Florida Institute of Technology
150 W. University Blvd
Melbourne, Fl 32901
nfischer2018@my.fit.edu

Ryan T. White
Florida Institute of Technology
150 W. University Blvd
Melbourne, Fl 32901
rwhite@my.fit.edu

Markus Wilde
Florida Institute of Technology
150 W. University Blvd
Melbourne, Fl 32901
mwilde@fit.edu

Brian Kish
Florida Institute of Technology
150 W. University Blvd
Melbourne, Fl 32901
bkish@fit.edu

Isaac Silver
Energy Management Aerospace
2000 General Aviation Drive
Hangar 101
Melbourne, FL
isaac@energymanagementaero.com

*Abstract*— Autonomous navigation and path-planning around non-cooperative space objects is an enabling technology for on-orbit servicing and space debris removal systems. The navigation task includes the determination of target object motion, the identification of target object features suitable for grasping, and the identification of collision hazards and other keep-out zones. Given this knowledge, chaser spacecraft can be guided towards capture locations without damaging the target object or without unduly the operations of a servicing target by covering up solar arrays or communication antennas. One way to autonomously achieve target identification, characterization and feature recognition is by use of artificial intelligence algorithms. This paper discusses how the combination of cameras and machine learning algorithms can achieve the relative navigation task. The performance of two deep learning-based object detection algorithms, Faster Region-based Convolutional Neural Networks (R-CNN) and You Only Look Once (YOLOv5), is tested using experimental data obtained in formation flight simulations in the ORION Lab at Florida Institute of Technology. The simulation scenarios vary the yaw motion of the target object, the chaser approach trajectory, and the lighting conditions in order to test the algorithms in a wide range of realistic and performance limiting situations. The data analyzed include the mean average precision metrics in order to compare the performance of the object detectors. The paper discusses the path to implementing the feature recognition algorithms and towards integrating them into the spacecraft Guidance Navigation and Control system.

TABLE OF CONTENTS

## 1. INTRODUCTION

With the rapid proliferation of satellites in LEO and the existing space debris, the risk of space debris collision is increasing. This risk demands the need for Active Space Debris Removal Technology (ADR) and On-Orbit Servicing (OOS) for safer spaceflight environment.

OOS and ADR systems must be able to engage non-cooperative targets that are not equipped with operational aids such as navigation equipment or capture interfaces. These resident space objects (RSOs) could be tumbling at a substantial rate, be damaged structurally, or have uncontrolled extended appendages. These uncertainties in rendezvous and capture create a need for autonomous identification of non-cooperative target components.

OOS and ADR have been pursued since the start of spaceflight [1]. Previous concepts and technological developments include astronaut maintenance, pressurized "dry docks", and robotic refueling. During the Space Shuttle program, multiple missions of capturing and servicing or assembly (Palapa-B2, Westar VI, Hubble Space Telescope,

ISS [2]) shed light on the advantages of OOS. Because there is an associated risk with crewed OOS missions, crewed servicing progressed into autonomous robot servicing in the 1990s. After Japan's revolutionary ETS-VII mission in 1997 [3], NASA, DARPA, and AFRL presented OOS abilities with DART [4], XSS-10 [5], XSS-11 [6], ANGELS [7], and Orbital Express [8]. In 2007, Orbital Express autonomously captured, inspected, fueled, and swapped equipment. Following Orbital Express, NASA and DARPA plan to autonomously refuel a target in LEO (Restore-L Mission [9]), and service a Geostationary Equatorial Orbit (GEO) target (RSGS [10]) in 2022. Northrop Grumman launched Mission Extension Vehicles (MEVs) to upkeep GEO spacecraft [11, 12], marking the first commercial OOS service.

All previous autonomous missions, from ETS-VII to Orbital Express, serviced target spacecraft that were smaller than the specialized robot servicer. Each target was equipped with navigational aid and capture interfaces and maintained a stable attitude during capture and service. Upcoming missions such as Restore-L and RSGS will demonstrate OOS capabilities on non-cooperative targets.

In general, the ability to safely identify, approach, capture, inspect, and service an uncooperative vehicle has not been achieved. A non-cooperative target object could potentially be tumbling. This tumbling motion, in combination with the presence of a spacecraft's extended antennas or solar arrays, produces a very hazardous approach and capture environment. If a chaser satellite captures a target in motion, the target's structural hard points could have high motion rates relative to the chaser. Ultimately, this could result in severe damage due to high loads on both the chaser and target, leading to the generation of additional space debris.

To mitigate this problem, the use of autonomy and artificial intelligence are required. Machine learning algorithms, such as those discussed in this paper, can assist in in-space characterization, identification, and flight-planning.

The research reported in this paper contributes to this mission by training, testing, and comparing two machine learning algorithms; faster region-based convolutional neural network (Faster R-CNN) and You Only Look Once V5 (YOLOv5). Both algorithms identify and characterize target spacecraft bodies and solar arrays. However, each holds unique advantages as they vary in development and operation. This paper searches to determine which algorithm outperforms the other and provides the highest accuracy and efficiency of real-time component identification.

As such, this paper is following up on the work reported in [13, 14]. The research in both these papers uses the Ultralytics implementation of YOLOv5 to enable autonomous satellite feature recognition. The comparison study between YOLOv5 and Faster R-CNN performed in this paper studies the trade-off between detection accuracies and inference time for unique test cases.

The paper is structured as follows. Section 2 provides a literature review in machine learning algorithms with focus on YOLO and Faster R-CNN. Section 3 describes the dataset compiled for this research, which is used in algorithm training in Section 4 and for performance testing in Section 5. Section 6 discusses the implementation on a combination of Raspberry Pi 4 and Neural Compute Stick. Section 7 concludes the paper.

## 2. YOLO AND FASTER R-CNN

Computer vision is a discipline focused on automatic extraction of information from visual data with a history reaching back to the 1960s, emerging with grand aims to replicate the human visual system to contribute to the development of a generalized intelligence. Early researchers were primarily armed with hand-crafted filters for performing human-directed tasks, such as sharpening images, detecting edges and shapes, and blurring. These tasks were often carried out by allowing a small image kernel (e.g. a 5-by-5 matrix with carefully-selected elements) to scan the image, multiplying elementwise over each patch of pixels of the same size and summing to generate a feature map, i.e. a picture with pixel intensities for an image with enhanced edges, for example.

Parallel with classical computer vision (and with some crossover) was the development of models of biological neurons based on an early understanding of how animal brains work. These artificial neural neurons would observe input data (stimuli) and produce a binary output computed using some internal weights. These artificial neurons could solve linear classification problems on low-dimensional data. The weights were hand-crafted to solve the given problem until Rosenblatt's perceptron algorithm [15] allowed computers to automatically learn the weights to solve the problem. Building layers of interconnected perceptrons with weighted connections formed a multilayer perceptron, or (artificial) neural network. However, this was a far too computationally expensive approach for high-dimensional data, e.g., images, and it was restricted to solving linearly separable problems, a restriction that rendered the models useless for most computer vision tasks.

In the 1980s, new developments allowed use of neural networks for computer vision tasks. Nonlinear activation functions allowed neural networks to solve nonlinear classification problems and the backpropagation algorithm [16, 17] allowed for efficient computation of the gradient vector of a loss function with respect to the weights in the network to fit the model. This made gradient descent a viable approach to quickly train neural networks to minimize a loss function. LeCun et al. [18] drew on models of the mammalian visual cortex by Hubel and Wiesel [19] and a new neural architecture by Fukushima [20] to propose an approach unifying neural networks with classical computer vision: the convolutional neural network (CNN). They re-designed conventional neural architectures to use new convolutional layers that fed data forward by deploying several image kernels from classical computer vision to scan the images to extract features, but these image kernels were made up of learnable weights mapped by a differential function to the loss function. These convolutional layers, alternating with subsampling (max-pooling) layers, permitted the image kernel weights to be learned by backpropagation, a second time-decrease to model training with the capability to extract local features, improving performance in image classification.

*From Image Classification to Object Detection*

In 2012, Krizhevsky et al. [21] developed AlexNet, a larger version of LeCun's model with some innovations (e.g. ReLU activation), on graphics processing units (GPUs). Their unparalleled capabilities to parallelize the many arithmetic operations inherent in neural network processing and backpropagation permitted much deeper CNNs (i.e. with many more layers) with more weights and greater capacity to learn. Until 2014, deeper but similar CNNs, notably VGG Net [22], incrementally improved image classification performance.

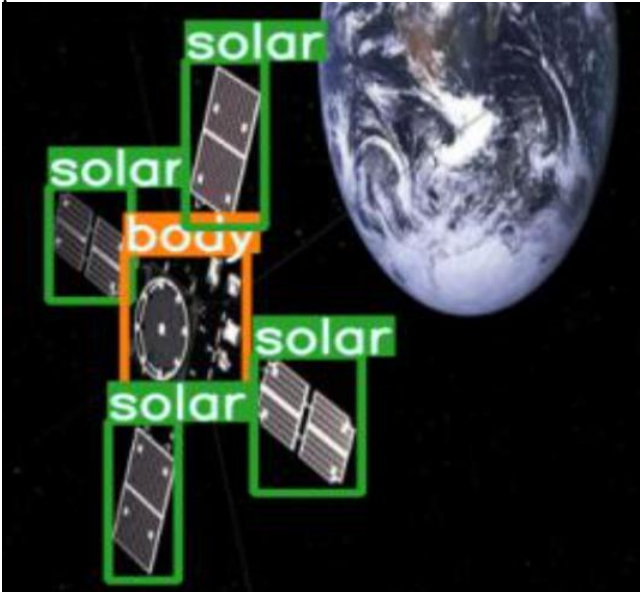

**Figure 1: A satellite with four solar arrays and a body, including bounding boxes and labels**

In 2014, Szegedy et. al. [23] introduced the Inception network, which effectively implemented a network-in-network model [24] that creates sub-networks between layers that would deploy learnable image kernels of various sizes—each adept at locating features at different scales. This tremendously reduced the number of parameters and improved training time and performance. In 2015, He et al. [25] introduced residual networks (ResNets), which add "skip-connections" that feed data forward by several layers at once, skipping the intermediate layers. This helps maintain the flow of information (gradients) through deep networks, permitting neural networks over 150 layers deep (nearly a 7-fold increase from Inception). This model outperformed humans at image classification on the popular benchmark dataset ImageNet [26].

With such tremendous performance on image classification, the computer vision research community moved on to more difficult tasks. In particular, *object detection* is a significantly more difficult task, which simultaneously classifies multiple objects in a single image and localizes each object by drawing a bounding box tightly surrounding it, as in Figure 1.

If an object detector will detect all satellite bodies and solar panels, it must predict an unknown number of bounding boxes, each including the center of the box ($x$, $y$) and the dimensions of the box $w$ and $h$, an objectness score $o$ corresponding to the posterior probability the box contains an object, and a posterior probability distribution $\mathbf{p} = (p_b, p_s)$ that the box contains a body or solar array. This makes for an 8-dimensional target: ($o$, $x$, $y$, $w$, $h$, $p_b$, $p_s$)

*Multi-Stage and Single-Stage Object Detection Algorithms*

While there are non-neural object detection algorithms, such as the feature extraction-based histogram of oriented gradients [27] paired with an SVM and GPU-accelerated CNN model OverFeat [28], the most effective objection detection algorithms at this time fall into two general categories: multi-stage detectors and single-stage detectors. Multi-stage detectors take an input image and split the object detection task into several discrete parts, ultimately outputting bounding boxes while single-stage detectors directly map input images to bounding box predictions. In general, state-of-the-art multi-stage object detectors perform with the most accuracy but are more computationally expensive at inference time than single-stage object detectors. There is a trade-off between the high accuracy of multi-stage detectors and high framerate of single-stage detectors when attempting to run real-time object detection on the frames of a video or camera feed.

The region-based CNN (R-CNN) is a multi-stage object detector developed by Girschick et al. [29] that works by breaking object detection into four tasks: propose regions where objects may exist, use a CNN to extract features from each region proposal, classify each proposed region with a support vector machine (SVM), and use a regression model to predict bounding boxes. Rather than proposing every sub-region of an image for classification, R-CNN greedily combines similar regions proposals into a smaller number of region proposals (from a few hundred to a few thousand) with a selective search algorithm. Unfortunately, selective search is expensive to run and, even after that, running CNN inference on hundreds of images is expensive, making real-time infeasible with an R-CNN.

Fast R-CNN [30] addresses some of the shortcomings of R-CNN. It re-orders the first two stages of the object detection task and merges the last two tasks. Now, the entire image is fed into a CNN, mapping it to a lower-dimensional feature map, and then selective search generates region proposals. These region proposals are then fed into a new region of interest pooling layer that maps all regions to the same dimensions so they can be fed into a small, fully connected neural network to simultaneously predict bounding boxes and classify the objects inside. Fast RCNN has as much as a twenty-fold framerate improvement over RCNN because its CNN only has to process one image rather than thousands, region proposals come from a lower dimensional feature map, and the prediction is combined.

Despite the computational improvement, Fast R-CNN still takes multiple seconds on conventional hardware to perform object detection on a single image implying framerates below 1 Hz with desktop hardware. Its successor, Faster R-CNN by Ren, et al. [31], throws out selective search entirely and replaces it with a learnable region proposal network (RPN), a CNN that can be trained such that it simply needs to run inference to enable object detection, bringing the framerate as high as 17 Hz with conventional hardware, which suggests Faster R-CNN is an option for real-time object detection from a camera feed. However, in the on-board application that we consider, it could not be a standalone solution due to computational constraints but may be feasible to use sparsely in combination with a faster algorithm to fill in the gaps between Faster R-CNN inferences.

Single-stage object detectors propose a radically different approach: training a single neural network to map an image to objectness scores, bonding boxes, and object classifications. The Single Shot Detector (SSD) by Liu et al. [32] throws out region proposals entirely to dramatically improve object detection framerates at a cost of some accuracy compared to Faster R-CNN.

About the same time, the You Only Look Once (YOLO) object detector was developed by Redmon et al. [33], which runs much more quickly than even SSDs. The general idea of YOLO is to partition the input image into cells with a grid, efficiently predict several bounding boxes centered in each grid-cell along with objectness scores and classifications with an Inception-like CNN called DarkNet so it can learn features at different scales. YOLO will then remove boxes with low objectness scores and uses a method called non-maximum suppression to choose the best remaining box based on the intersection over union (IoU) similarity metric. This is a tremendously cheaper way to select the best box than the region proposals seen in the R-CNN family. Though very fast at inference time, YOLO lacks the accuracy of Faster R-CNN.

A later refinement YOLO9000 (also known as YOLOv2) [34] provided several improvements resulting in higher accuracy, notably the introduction of *anchor boxes* predetermined by K-means clustering to serve as starting points for bounding-box predictions by learning typical sizes and aspect ratios for bounding boxes from each object class. This allows YOLOv2 to predict good bounding boxes more effectively in less time. YOLOv3 [35] is an incrementally improvement that modifies the loss function and implements ResNet-type skip-connections, resulting in greater accuracy at no computational cost at inference time.

Bochkovskiy et al. [36] introduced YOLOv4, less a new algorithm and more a wide-ranging study of methods to optimize YOLO-style object detectors. The authors provide extensive hyperparameter-tuning experiments, resulting in several best practices for choosing the CNN architecture, activation functions, loss functions, and methods for data augmentation, regularization, and normalization. Further, it provides a comparison of hyperparameters related to the training algorithm. The resulting YOLOv4 implementation gains some important but incremental improvements in training time, accuracy, and inference time.

Jocher et al. [37] released a convenient PyTorch implementation of YOLO entitled YOLOv5. It is a full-featured implementation making several innovations and tremendous quality-of-life improvements for practitioners to permit easy training and tuning of the object detector, including a genetic algorithm for searching the hyperparameter space, integrations with popular data annotation tools like Roboflow [38], and many of the optional features suggested by the YOLOv4 paper. Despite some controversy between the developers of YOLOv4 and YOLOv5, they seem to perform similarly in both accuracy and runtime by all indications, but we found the latter to be more convenient to use.

As the state-of-the-art multi-stage and single-stage object detectors, respectively, the present work compares the performance of Faster R-CNN to YOLOv5 for detecting solar arrays and bodies of satellites. The literature indicates Faster R-CNN should be more accurate, but YOLOv5 should run inference more quickly, but the specific balance of these metrics is unclear for our application.

It should be noted that training both YOLO and R-CNN, or any deep neural network, requires strong computational power and cannot be done on-board. This training performed on the ground learns model parameters that enable the trained models to be deployed to perform inference alone with the on-board hardware. This supports the wider goal of developing chaser satellites that can autonomously plan real-time flightpaths to the bodies of uncooperative satellites avoiding the sensitive components such as solar panels, antennas etc., based solely on in-flight camera feeds and decisions made on-board.

The authors are unaware of prior usage of YOLOv5 or Faster R-CNN in space, but there are several papers proposing the use of modern object detection algorithms for detecting satellite components. Our team proposed YOLOv5 in prior work [13, 14] trained on a mixture of synthetic and real RGB images and tested on images captured at the ORION Lab [39] at Florida Tech. The work is unique in aiming to perform object detection on-board, using Faster R-CNN and YOLOv5 for satellite component detection, and performing hardware-in-the-loop testing for this problem.

Object detectors YOLOv3, YOLOv4, and EfficientDet were trained on a similar mixture of synthetic and real RGB images by Dung et. al. [40]. Advanced neural computer vision methods have been proposed in two related but different problems. First, Musallam et. al. [41] proposed a challenge in 2021 to detect entire spacecraft in RGB-depth imagery where some participants used YOLO on their dataset generated in the Unity3D game engine. Second, there are numerous works on spacecraft pose estimation centered around the ESA's Pose Estimation Challenge [42], where most entrants used neural networks trained on 15000 synthetic images and 300 real images captured at the Space Rendezvous Laboratory (SLAB) at Stanford University.

### 3. DATASET

Each object detection algorithm is trained on a training dataset and tested on both validation and testing datasets. For this work, the training data consists of images of satellites from Google image search and NASA archived images.

The dataset was developed based on certain criteria [13]:

1. The images must contain at least one of the spacecraft features of interest (solar panels, and satellite bodies).
2. The features in the images must be recognizable by humans.
3. The features must be relatable to real-life satellite components.
4. No image may be repeated.

Based on these criteria, a total of 1231 RGB images were gathered for training. These images were then annotated using Roboflow[1] with bounding boxes and labels for all solar arrays and satellite bodies. Figure 3 is an image of the annotation heatmap of the image dataset. The heatmap shows that most of the features lie in the center of the image. This information among other criteria suggests which data augmentation techniques to use while training the object detector.

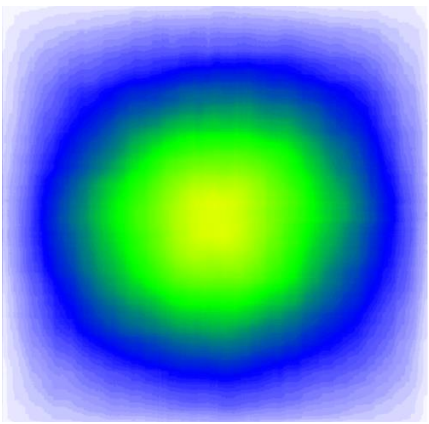

**Figure 3: Annotations Heatmap**

The dataset is split randomly with 71% in training data, 25% in validation data, and 4%. in testing data. Figure 2 shows an example of the raw data image set. Each colored box within the image represents a feature.

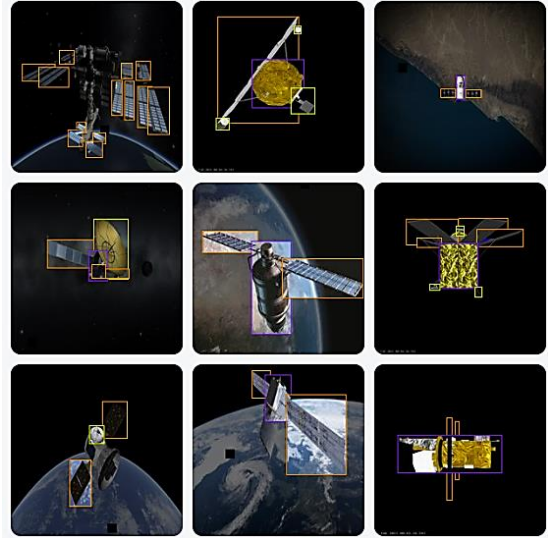

**Figure 2: Raw Dataset Example**

IR, near UV, LIDAR, and thermal imagery was not used in the training dataset, as it was unavailable in suitable quantities to train a deep neural network without resorting to a less visually diverse dataset that could be captured with simulated imagery. Multi-modal videos involving RGB and depth channels captured with an Intel RealSense stereographic camera has been captured in the ORION Lab. Prior work uses depth for on-board flightpath planning based on the real-time object detections delivered by YOLOv5 [14]. The present work explores ways to improve the object detection, so only RGB videos are used in testing in Section 4.

In this paper, two algorithms, YOLOv5 and Faster R-CNN, are evaluated on experimental test data. The experimental test data consists of real-life videos of a target spacecraft in an approaching chaser's point of view. These videos were recorded at the ORION research laboratory at the Florida Institute of Technology. The ORION lab is equipped with a hardware-in-the-loop formation flight and docking simulator and, a Hilio Litepanel D12 350W LED light fixture with a 5,600 K color temperature which is used to create orbital lighting effects [39]. A total of four video cases were studied in this paper. Figure 4 summarize the test cases. The experimental image dataset was then created by extracting one video frame per second. The models, however, will not see the ground mockup during training, so we demonstrate performance of the algorithms at detecting features on a satellite it has never seen.

---

[1] https://roboflow.com

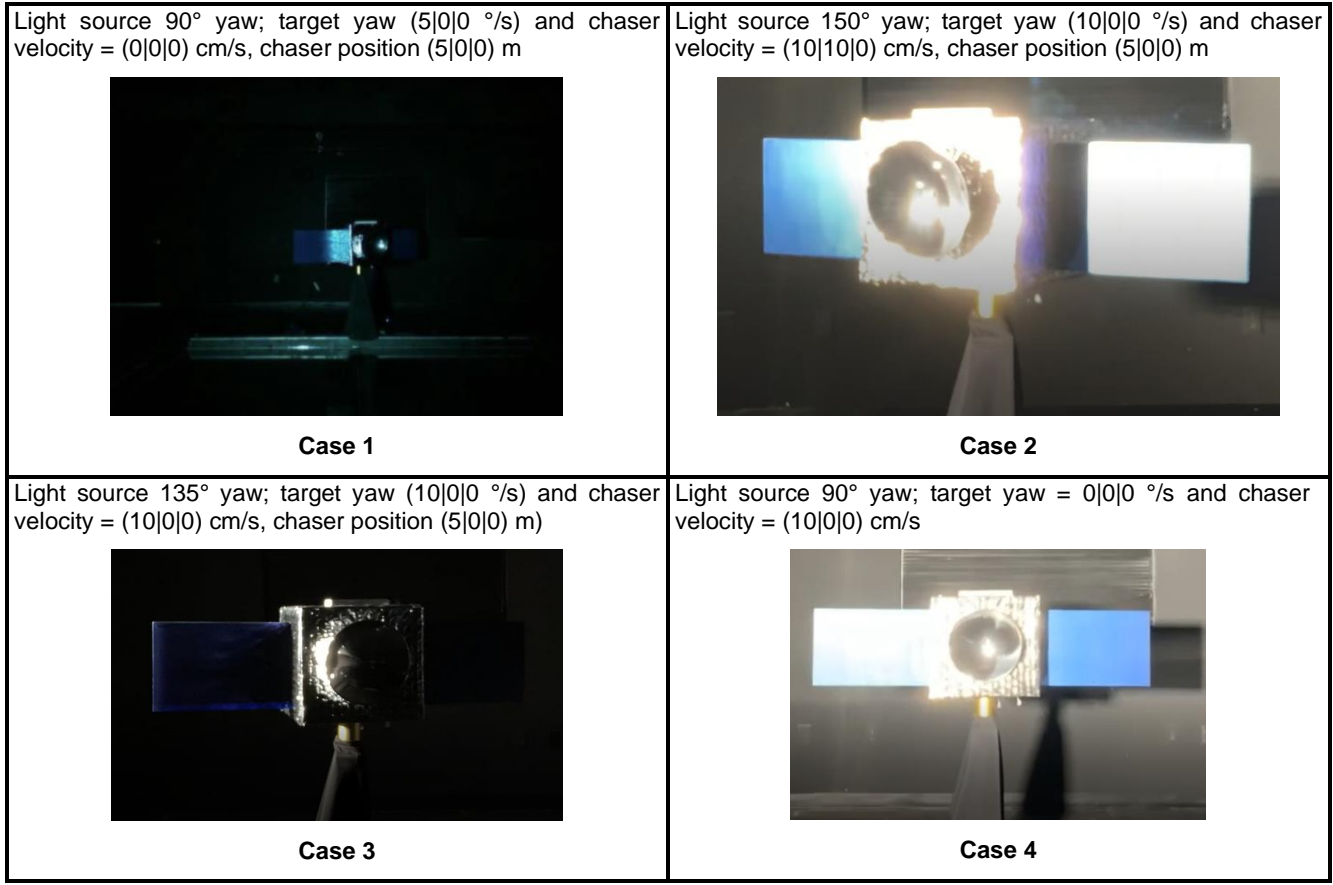

**Figure 4: Test Cases**

## 4. TRAINING

The following augmentations were applied:

1. Random 90° rotation in either the clockwise or counter-clockwise direction
2. Random rotation by either -45° or 45°
3. Random application of grayscale to 25% of total images
4. Random application of Gaussian blur up to 2 pixels
5. Random adjustment of bounding box brightness between -31% and 31%
6. Random adjustment of bounding box exposure between -34% and 34%

These augmentations were chosen based on issues such as exposure, brightness, noise, blur the Intel RealSense D435i camera experienced while recording experimental videos for the experiment conducted in [14].

The Detectron2 implementation of Faster R-CNN model was implemented in PyTorch [43]. It uses transfer learning for training, starting with the resnet101 model pre-trained on the COCO dataset. Figure 5 - Figure 8 show plots of performance metrics (vertical axis) over the number of epochs (horizontal axis) that are used to compare YOLOv5 and Faster R-CNN

Note that YOLOv5 model was trained using the best pre-trained weights obtained after every 100 epochs for 4 times. The total number of training epochs in the plots below only represent the last 100 epochs the algorithm was ran for. This does not affect the metrics obtained from training the algorithm, however.

*Mean Average Precision - mAP*

mAP is the mean average precision over all classes in the dataset:

$$mAP = \frac{1}{|classes|} \Sigma_{c \epsilon classes} \# \frac{TP(c)}{\#TP(c) + \#FP(c)} \quad (1)$$

mAP@0.5 represents mean average precision of predictions with 50% Intersection over Union (IoU). Similarly, mAP@0.5:0.95 5 represents mean average precision of predictions with IoU between 50% and 95%. Hence, the higher the mAP at higher IoU, the more accurate the algorithms are.

*Training Classification Loss*

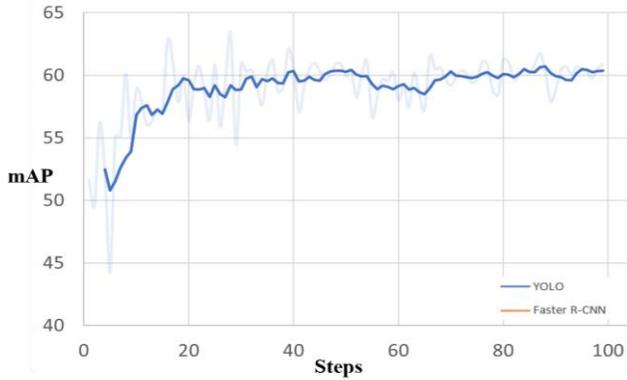

Figure 5: YOLO mAP@0.5 IoU

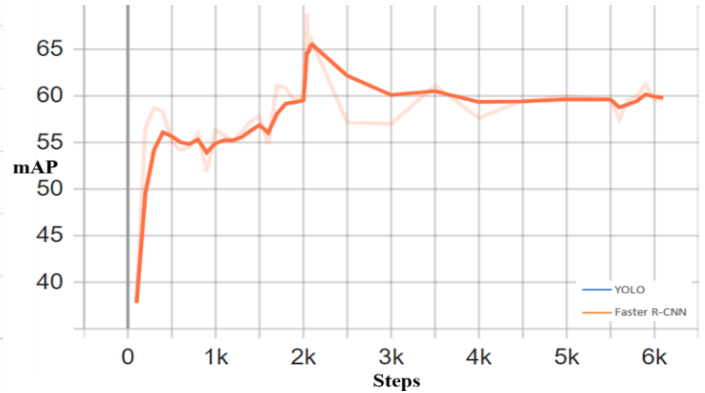

Figure 6: Faster R-CNN mAP@0.5 IoU

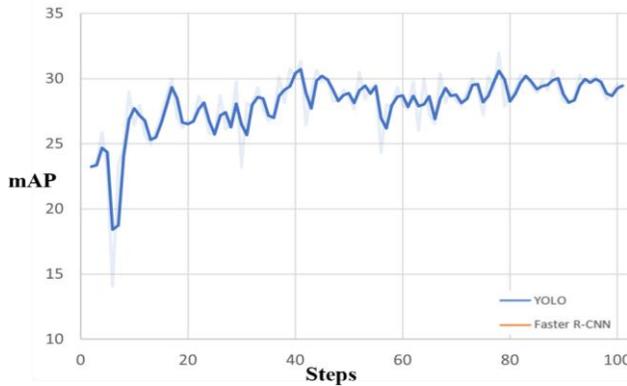

Figure 7: YOLO mAP@0.5:0.95 IoU

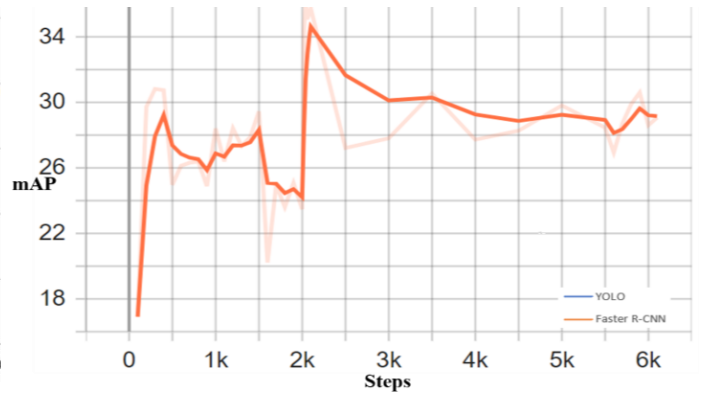

Figure 8: Faster R-CNN mAP@0.5:0.95 IoU

Intersection over Union is defined as the ratio of overlapping area between the prediction bounding box and the ground truth box divided by the union of both the boxes.

The sudden drop/increase in the trend of the plots at epoch 2000 in Figure 7, and Figure 8 is due to re-starting training with pre-trained weights for Faster R-CNN.

The trend in the figures above show both YOLOv5 and Faster R-CNN have reached a saturation point with the initialized hyperparameters. Furthermore, Faster R-CNN and YOLOv5 seem to saturate around the same values.

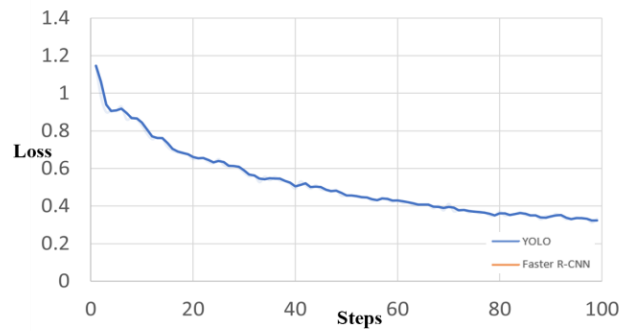

Figure 9 and Figure 10 show the training epoch vs. classification loss function. The goal of both the algorithms is to minimize the classification loss to achieve more accurate detections

Finally, detection weights corresponding to the best training and testing metrics were obtained for testing the two algorithms.

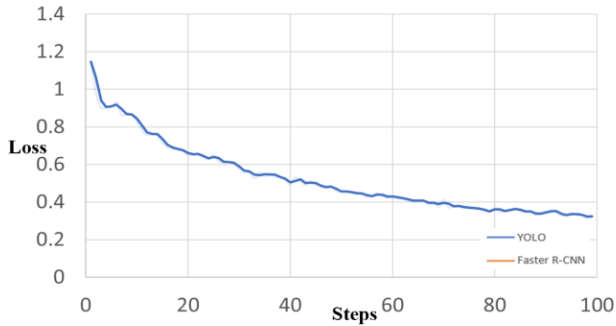

**Figure 9: YOLO Classification Loss**

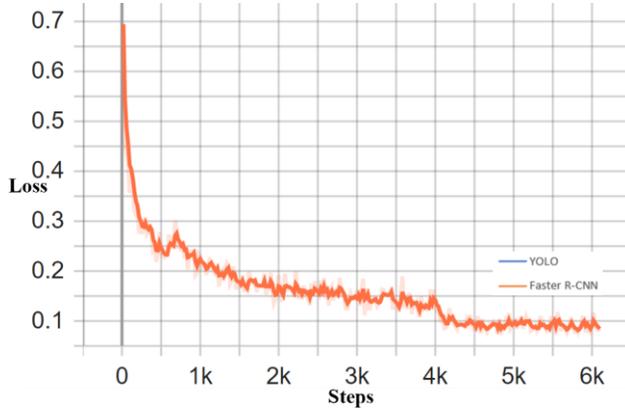

**Figure 10: Faster R-CNN Classification Loss**

### 5. TESTING

The two algorithms use mean Average Precision (mAP) at Intersection over Union of 0.5 and 0.5:0.95 as well as average recall values as the common testing metrics [44].

Based on the weights obtained from training the algorithm, the model was deployed on videos of the 4 experimental cases (see Figure 4) captured at the ORION Lab outlined in Section 3. These are videos of a satellite the algorithms have never seen in training. Results for each test case are tabulated below.

*Case 1*

**Table 1: YOLO Case 1 Results**

| Class | Images | Targets | AP | AR | mAP@.5 | mAP@.5:.95 |
|---|---|---|---|---|---|---|
| All | 81 | 214 | 88.4% | 70.8% | 75.4% | 34.5% |
| Body | 81 | 81 | 89.5% | 84% | 90.2% | 47.2% |
| Solar | 81 | 133 | 87.4% | 57.6% | 60.6% | 21.8% |
| Inference Rate: | 0.017 s/img | | | | | |

**Table 2: Faster R-CNN Case 1 Results**

| Class | Images | Targets | AP | AR | mAP@.5 | mAP@.5:.95 |
|---|---|---|---|---|---|---|
| All | 81 | 211 | | 16.7% | 32% | 12.6% |
| Body | 81 | 80 | | | | 33.97% |
| Solar | 81 | 131 | | | | 13.84% |
| Inference Rate: | 0.17s /img | | | | | |

*Case 2*

**Table 3: YOLOv5 Case 2 Results**

| Class | Images | Targets | AP | AR | mAP@.5 | mAP@.5:.95 |
|---|---|---|---|---|---|---|
| All | 108 | 310 | 38% | 39.2% | 32.1% | 19.9% |
| Body | 108 | 108 | 29.6% | 34.3% | 21.2% | 11.0% |
| Solar | 108 | 202 | 46.4% | 44.1% | 43% | 28.7% |
| Inference Rate: | 0.018 s/img | | | | | |

**Table 4: Faster R-CNN Case 2 Results**

| Class | Images | Targets | AP | AR | mAP@.5 | mAP@.5:.95 |
|---|---|---|---|---|---|---|
| All | 108 | 310 | | 49.8% | 61.2% | 37.7% |
| Body | 108 | 108 | | | | 58.552% |
| Solar | 108 | 202 | | | | 16.875% |
| Inference Rate: | 0.17s /img | | | | | |

*Case 3*

**Table 5: YOLOv5 Case 3 Results**

| Class | Images | Targets | AP | AR | mAP@.5 | mAP@.5:.95 |
|---|---|---|---|---|---|---|
| All | 34 | 103 | 51.3% | 41.4% | 40.5% | 21.3% |
| Body | 34 | 35 | 42.8% | 34.3% | 29.1% | 17.5% |
| Solar | 34 | 68 | 59.9% | 48.5% | 52% | 25.2% |
| Inference Rate: | 0.017 s/img | | | | | |

**Table 6: Faster R-CNN Case 3 Results**

| Class | Images | Targets* | AP | AR | mAP@.5 | mAP@.5:.95 |
|---|---|---|---|---|---|---|
| All | 34 | | | 57% | 70.5% | 40.1% |
| Body | 34 | | | | | 53.531% |
| Solar | 34 | | | | | 26.693% |
| Inference Rate: | 0.17s /img | | | | | |

*Data was not saved. File crashed

*Case 4*

**Table 7: YOLOv5 Case 4 Results**

| Class | Images | Targets | AP | AR | mAP@.5 | mAP@.5:.95 |
|---|---|---|---|---|---|---|
| All | 43 | 129 | 92.4% | 52.3% | 64.2% | 51.2% |
| Body | 43 | 43 | 100% | 48.7% | 56.2% | 48.2% |
| Solar | 43 | 86 | 84.8% | 55.8% | 71.6% | 54.2% |
| Inference Rate: | 0.017 s/img | | | | | |

**Table 8: Faster R-CNN Case 4 Results**

| Class | Images | Targets | AP | AR | mAP@.5 | mAP@.5:.95 |
|---|---|---|---|---|---|---|
| All | 43 | 129 | | 79.8% | 90.6% | 73.7% |
| Body | 43 | 43 | | | | 82.103% |
| Solar | 43 | 86 | | | | 65.347% |
| Inference Rate: | 0.17s /img | | | | | |

All the results from the Tables above are summarized in Table 9. It is seen that Faster R-CNN by far has better average mAP @ IoU 0.5, and 0.5:0.95 values compared to YOLOv5.

Upon analyzing the results for each test case, it is seen that lighting conditions play a major role in detection performance. Comparing mAP@0.5 and mAP@0.5:0.95, YOLOv5 performs better than Faster R-CNN for Case 1 with dimmer lighting with the light source placed at 90º yaw. This difference in performance suggests the training dataset has a gap with intense lighting, which may be improved by implementing lighting augmentations or gathering more source data with intense lighting to expand the training dataset so the algorithms learn to detect objects based on context when intense lighting obscures them. Another option is to pass the camera feed through a computationally efficient hand-crafted convolutional filters to soften the lighting and enhance edges before inference.

Additionally, YOLOv5 detects solar panels better than Faster R-CNN in video cases 1 and 2 and, almost the same as Faster R-CNN in video case 3. These three videos captured the target satellite rotating about its vertical axis causing the solar panels to disappear/hide behind the satellite body. These results prove the ability of YOLOv5 to detect features in the background accurately.

Overall, Faster R-CNN was a bit more accurate than YOLOv5; however, using Faster R-CNN comes with a cost of higher inference time. On average, Faster R-CNN's inference rate is 10 times slower than YOLOv5.

**Table 9: Summary of Experimental Results**

|  | YOLOv5 | | | Faster R-CNN | | |
|---|---|---|---|---|---|---|
| Class | AR | mAP@.5 | mAP@.5:.95 | AR | mAP@.5 | mAP@.5:.95 |
| All | 50.925% | 53.05% | 31.725% | 51.875% | 63.575% | 41.03% |
| Body | | | 30.957% | | | 57.03% |
| Solar | | | 32.475% | | | 30.69% |
| | Inference Rate: 0.017s/img | | | Inference Rate: 0.17 s/img | | |

To find inference rates for real-time detection, both the algorithms were ran in CPU mode on a laptop with an Intel I7 8550U with an input video camera. YOLO inferred at 0.471 s/frame while, Faster R-CNN inferred at 4.12s/frame.

For this research, time is a crucial aspect of the mission. Lower inference rate means lesser frame rate. The satellite would not receive enough updates on the RSO's position and orientation. This would lead to unsuccessful missions as the chaser might miss or even worse, collide with the target. Thus, YOLOv5 is preferred over Faster R-CNN. However, having Faster R-CNN's detections output at spaced out intervals while YOLO continues inferring would be a plausible way to verify YOLO's output. Faster R-CNN could also be used to infer when the chaser and target are not in close proximity.

## 6. IMPLEMENTATION

The models discussed in this paper are not natively capable of running efficiently on low powered hardware. They must be modified for computing power that is similar to hardware that is designed for space operation. Currently in development, the High-performance spaceflight computing program has a design which features 8 ARM Cortex-A53 cores running at 800Mhz [45]. This would provide similar performance to the Raspberry Pi model 4B which has 4 ARM Cortex-A72 cores running at 1500Mhz [46].

Currently, these object detection models are modified using the OpenVINO toolkit to run on the Raspberry Pi. The OpenVINO toolkit optimizes the model for running on Intel hardware, the hardware for this project is the Intel Neural Compute Stick 2 (NCS2). The NCS2 contains the Movidius Myriad X VPU or Vision Processing Unit, which provides hardware acceleration for deep learning tasks [47]. The NCS2 connects to the Raspberry Pi using USB.

To perform experimental verification of this implementation, the YOLOv5 model was converted to OpenVINO and ran per GitHub repository [48]. Though not experimentally verified by this research yet, OpenVINO also has support for Faster R-CNN [49].

Our future goals involve implementation of both the algorithms on separate Raspberry Pi's to perform detections in parallel. This study will aim at enhancing performance by supplementing high-framerate YOLOv5 detection by with less frequent, more accurate Faster R-CNN detections.

## 7. CONCLUSION

With the increasing interest in LEO missions and the risk of space debris collisions, OOS and ADR become a priority. This paper performs additional research on state-of-the-art deep learning-based object detectors to aid autonomous machine vision based OOS operation. The two object detectors studied in this paper are the current best single stage detector - YOLOv5 and the current best multi-stage detector – Faster R-CNN. The training data for both algorithms consist of spacecraft images found on google and NASA's archived images. Interested features such as solar panels and satellite body were labelled for each image. For the testing dataset, four videos of a target satellite under different lighting conditions and initial rotations were shot by a chaser spacecraft with a specific approach path. These videos were then segregated into images at a data rate of 1 frame per second and labelled for the same features as the training dataset.

Per certain initial hyperparameter conditions, both the algorithms were trained, and best weights corresponding to the best metrics were obtained. These were used for inferring the testing dataset for the classes solar panels and satellite body. The results from the testing dataset concluded that Faster R-CNN performs better than YOLOv5 however, the inference rate of YOLOv5 is 10 times higher than Faster R-CNN making YOLOv5 the preferred real-time object detector for this application.

Future work for this research includes running both YOLOv5 and Faster R-CNN in parallel to counterbalance lower mAP values of YOLOv5 with higher mAP values of Faster R-CNN. Additionally, more classes such as antennas and thruster nozzles along with different augmentation techniques will implemented to study the performance of both the algorithms.

## ACKNOWLEDGEMENTS

The work on this study was supported by AFWERX STTR Phase II contract FA864921P1506 and NVIDIA Applied Research Accelerator Program.

## BIOGRAPHY

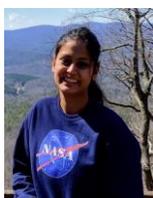 ***Trupti Mahendrakar*** received a B.S. in Engineering from Embry-Riddle Aeronautical University, Prescott, Arizona in 2019. She is currently a PhD student at Florida Institute of Technology, Melbourne. Her current research includes implementation of machine vision algorithms to enhance on-orbit service satellite operations, optimization of cold gas thruster design, implementation of a robotic arm for satellite refueling.

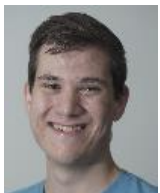 ***Andrew Ekblad*** is an Electrical Engineering senior at Florida Institute of technology. His interests include machine learning, photonics, mathematics, computer architecture, and signal Processing.

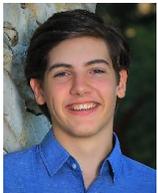 ***Nathan Fischer*** is currently an undergraduate student acquiring a BS degree in Aerospace Engineering from Florida Institute of Technology. He works as an artificial intelligence and control systems researcher in the Orbital Robotics Interaction, On-orbit servicing, and Navigation (ORION) laboratory. His current research surrounds satellite component recognition and rendezvous and capturing flight planning. He also leads an aerospace design team that is currently developing a compact and agile propulsion system that uses solid propellant and electrical ignition.

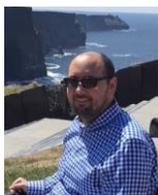 ***Ryan T. White*** earned his PhD from Florida Institute of Technology in 2015. He was appointed Assistant Professor in 2019 at the university. He focuses on multidisciplinary projects involving deep learning, computer vision, machine learning, and statistical models, including projects in-orbit satellite component detection, tracking facial drooping in patients with neurological disorders, and tracking glaciers from satellite imagery as well as machine learning work on marine biology and sustainable global development. His background and some continued work focus on probability and stochastic analysis.

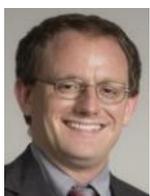 ***Dr. Markus Wilde*** is an Associate Professor for Aerospace Engineering at Florida Tech and Director of the ORION Lab. His research focus lies on experimental studies of spacecraft and aircraft control systems. Dr. Wilde received his M.S. and Ph.D. in Aerospace Engineering at TU Munich, Germany. He was accepted into the NRC Research Associateship Program in 2013, as postdoctoral associate at the Spacecraft Robotics Laboratory at the Naval Postgraduate School. In 2014, he joined the Florida Tech faculty.

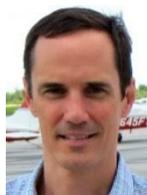 ***Dr. Brian Kish*** is the Chair of Florida Tech's Flight Test Engineering Program. He earned a Ph.D. in Aeronautical Engineering from the Air Force Institute of Technology. He is a graduate of the Air Force Test Pilot School and has accumulated over 1300 flight hours as a Flight Test Engineer in 49 different aircraft during his 20-year Air Force career. He held leadership positions at three Flight Test units and served as the Vice Chair of the Education Department of the Air Force Test Pilot School from 2005-2008. Since retiring from the Air Force in 2011, Dr. Kish has taught Control Systems, Aircraft Stability & Control, and Avionics courses at Florida Tech.

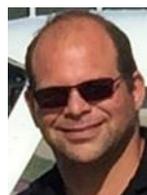 ***Dr. Isaac Silver*** is the CEO of Energy Management Aerospace. He earned his Ph.D. from Florida Tech in Space Sciences. He also holds a B.S. in Astronomy and Astrophysics from Florida Tech. He's an Airline Transport Pilot (Airplane Multi-Engine Land), Commercial Pilot (Airplane Single-Engine Land and Sea), Gold Seal Flight Instructor (Single and Multi-Engine), Instrument Pilot (Airplane) and Ground Instructor (Advanced). He has 17,500 hours of flight time with more than 4,000 hours as an instructor. Aircraft include DA-10, DA-20, DA-50/900, L-1329, BE-350, BE-400, IAI-1124, Learjet (24/25/31/35), and L39.